\pdfoutput=1
\documentclass{article}

\usepackage{ijcai24}

\usepackage{times}
\usepackage{soul}
\usepackage{url}
\usepackage[hidelinks]{hyperref}
\usepackage[utf8]{inputenc}
\usepackage[small]{caption}
\usepackage{graphicx}
\usepackage{amsmath}
\usepackage{amsthm}
\usepackage{booktabs}
\usepackage{algorithm}
\usepackage{algorithmic}
\usepackage[switch]{lineno}

\usepackage{multicol}
\usepackage{multirow}

\usepackage{threeparttable}

\usepackage{color}
\usepackage{multirow}

\title{Semi-Supervised Semantic Segmentation Based on Pseudo-Labels: A Survey}
\author{
Lingyan Ran$^1$ \and
Yali Li$^1$ \and
Guoqiang Liang$^{*1}$ \and
Yanning Zhang$^{1}$\\
\affiliations
$^1$School of Computer Science, Northwestern Polytechnical University\\
\emails
\{lran,gqliang, ynzhang\}@nwpu.edu.cn,
yarili@mail.nwpu.edu.cn
}

\begin{document}

\maketitle

\begin{abstract}
Semantic segmentation is an important and popular research area in computer vision that focuses on classifying pixels in an image based on their semantics. However, supervised deep learning requires large amounts of data to train models and the process of labeling images pixel by pixel is time-consuming and laborious. This review aims to provide a first comprehensive and organized overview of the state-of-the-art research results on pseudo-label methods in the field of semi-supervised semantic segmentation, which we categorize from different perspectives and present specific methods for specific application areas. In addition, we explore the application of pseudo-label technology in medical and remote-sensing image segmentation. Finally, we also propose some feasible future research directions to address the existing challenges.
\end{abstract}

\section{Introduction}
% 重要性 
% Deep learning was first combined with the field of computer vision by applying Convolutional Networks (CNNs) \cite{rawat2017deep} to the single image classification problem, obtaining excellent results. In recent years, it has been very closely linked to the field of computer vision, and many deep learning-based methods in computer vision have shown excellent results \cite{liu2022act}
% % liu2020deep,wang2020deep,minaee2021image，wang2021deep,
%, indicating its great potential for application in the field of computer vision. 
Semantic segmentation is an important and well-received research area in the field of computer vision to classify every pixel in an image, and it has a wide range of applications in specific areas such as medical image segmentation~\cite{li2023semi}
and remote sensing image segmentation \cite{wang2022satellite}. 
% In recent years, numerous successful approaches have been proposed, mainly benefiting from deep learning methods.
% Initially, segmentation primarily relied on conventional techniques such as clustering algorithms. %sezgin2004survey \cite{chen1998image}
Over the past few years, many works have made significant progress in improving the effectiveness of semantic segmentation tasks.
However, supervised deep learning requires large amounts of data to train models and the process of labeling images pixel by pixel is time-consuming and labor-intensive.
%, especially in specialized application domains, such as the medical field. 
Studies have pointed out that it takes hours to annotate just one finely labeled image from the Cityscapes dataset~\cite{cordts2016cityscapes}. 
The performance of fully supervised models may not be able to be improved significantly due to the cost of training labels.
%Thus, researchers have proposed the introduction of weakly supervised, 
% \cite{6247757}, 
%semi-supervised,%\cite{lazarova2014semi}, 
%and unsupervised learning.%\cite{comaniciu2002mean}.
% 
% 半监督语义分割介绍
%
In recent times, semi-supervised learning has been applied to semantic segmentation through numerous associated research studies.
% 
% 指出没有重点关注本文所讲的方法，突出本文的意义
% A review study \cite{pelaez2023survey} proposed a taxonomy that systematically classifies these methods 
% into pseudo-label,
% % li2021residual,teh2022gist,kim2022semi,zhang2022semi,\cite{yang2022st++}
% consistency regularization,  
% % french2019semi,ouali2020semi,ke2020guided,lee2021anti,\cite{wu2021semi}
% contrastive learning, 
% % alonso2021semi,liu2021bootstrapping,qiao2023fuzzy,wang2023space,\cite{xie2023boosting}
% adversarial training, 
% % souly2017semi,li2021semantic,zhang2021stable,jin2021adversarial,\cite{lee2022voxel}
% and hybrid methods.
% % lai2021semi,zhang2022region,xiao2022semi,\cite{hou2022semi}

\begin{figure}[!t]
    \centering
    \includegraphics[width=1\linewidth]{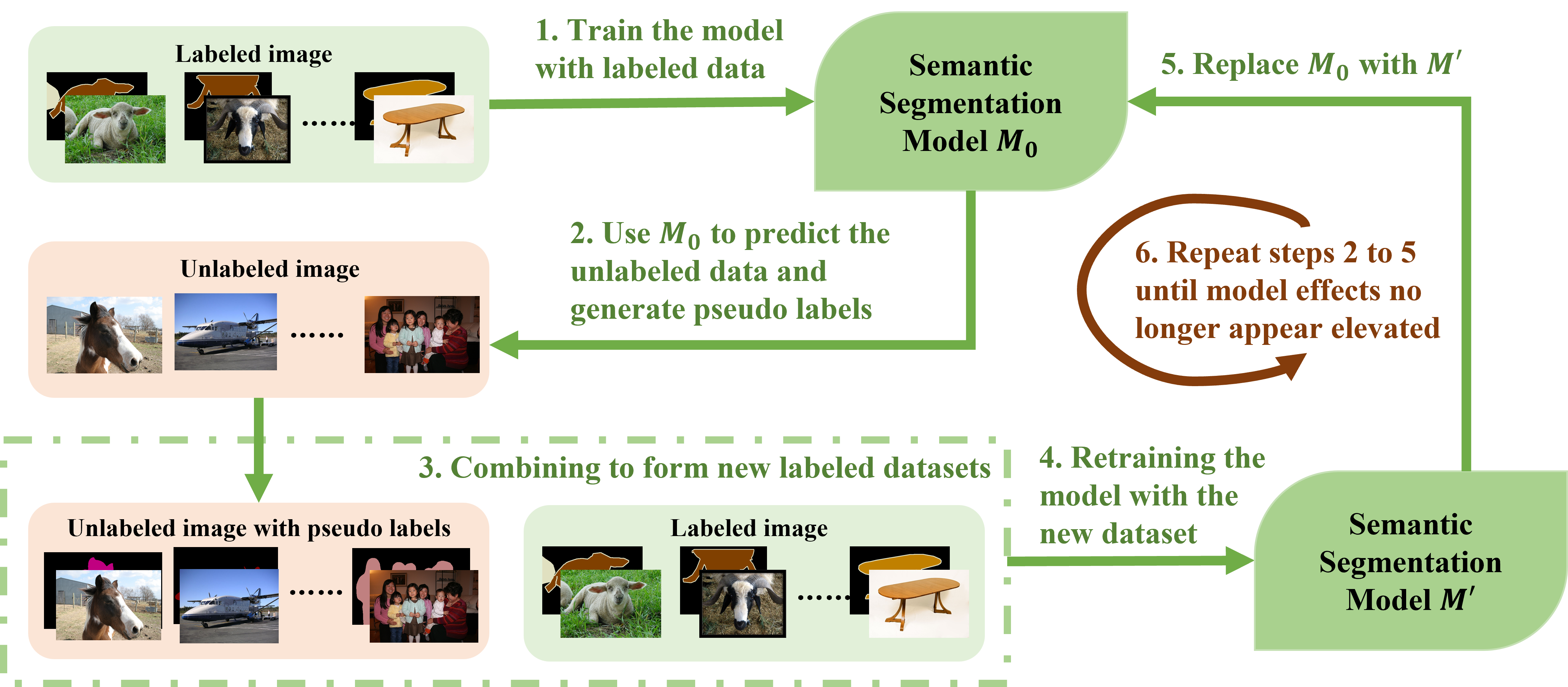}
    \caption{The plainest process of semi-supervised semantic segmentation with pseudo-label method. The acquisition and qualification of pseudo-labels are the main focus of the framework.}
    \label{fig_pross}
    % \setlength{\floatsep}{2pt plus 2pt minus 2pt}
    % The basic process is to generate pseudo-labels using models trained with labeled data. These pseudo-labels are utilized to retrain the model in conjunction with the labeled data.
\end{figure}

% 第一次提出伪标签 具体过程删掉
The pseudo-label method is a well-known technique in the semi-supervised learning field that first appeared in ~\cite{lee2013pseudo} and has gained popularity in recent computer vision research, including 
% image classification ~\cite{seydgar2022semisupervised},
domain adaptation ~\cite{li2023pseudo}, 
% object detection~\cite{hu2022pseudoprop},
semantic segmentation~\cite{wu2023querying},
etc., which are favored for their simplicity and impressive performance.
The process is depicted in Figure~\ref{fig_pross}. 
% 伪标签过程重复，删掉
% The basic concept of the pseudo-label is very simple: first, a model is trained using labeled data, then the model is used to generate pseudo-labels for unlabeled data, and finally, the model is retrained using a synthetic dataset. 
In semantic segmentation, the pseudo-label method is considered to be a more dependable option than consistent regularization, which may be affected by different levels of data augmentation. 
Pseudo-label technology is renowned for its stability, interpretability, and ease of implementation, making it an area of research with increasing potential.

% 写作动机
% 对比其他survey
Extensive research has been conducted on the use of the pseudo-label method in semi-supervised domains. Nevertheless, a current survey~\cite{pelaez2023survey} merely classifies semi-supervised semantic segmentation techniques systematically, lacking detailed summaries and analyses of pseudo-label methods. 
This deficiency has motivated us to undertake a survey. 

Our survey's primary objective is to offer a thorough and structured summary of recent research, categorizing the studies based on various perspectives, and presenting specific methods for specific application areas.
% 
% Overall, there are many researches based on the direction of pseudo-label in semantic segmentation in semi-supervised domains, but there is no work to statistically and analytically analyze these researches, which is the motivation for writing this paper to review and look forward to the pseudo-label methods in semi-supervised semantic segmentation. We provide a comprehensive and systematic overview of some of the recent works and rationally categorize them according to their different improvement perspectives.
% 
% 贡献
Our main contributions are:
1. In this paper, we provide a comprehensive review of the recent advancements in the pseudo-label methods for semi-supervised semantic segmentation. 
2. Specifically, we investigate three key aspects of the pseudo-label methods, which include the design of the model structure, refinement of pseudo-labels, and optimization techniques. 
3. Furthermore, we discuss the existing challenges in this field that require attention and propose potential directions for future research.

% 具体内容展开
% This paper is organized as follows: first, we introduce the background knowledge of the review contents in Section II; then, we analyze and review the existing semi-supervised semantic segmentation methods from five different perspectives specifically in Section III; then, we selectively analyze the current novel segmentation methods in Section IV; in Section V, we analyze the pseudo-label methods from the perspectives of quantitative experiments and qualitative evaluations; and, finally, we summarize the results in Section VI and propose the challenges that remain nowadays and the directions for future development.

% 主要聚集于伪标签方法
\section{Preliminary and Problem Formulation}
\subsubsection{Problem Definition}
In the context of semi-supervised semantic segmentation, the objective is to minimize the loss by considering both labeled and unlabeled datasets. The labeled dataset, denoted as ${D}_l={\{({x}_l,{y}_l)\}}^p$, consists of $p$ samples with corresponding labels. The unlabeled dataset, denoted as ${D}_u={\{{x}_u\}}^q$, consists of $q$ images, where $q$ is significantly larger than $p$. 
The loss function $L$ is defined as the sum of two terms: $L^l$ which represents the loss on the labeled dataset, and $L^u$ which represents the loss on the unlabeled dataset.
\begin{align}
    L = L^l + \lambda L^u,
\end{align}
where $\lambda$ represents a hyper-parameter that balances the trade-off. This hyper-parameter can either be assigned a fixed value beforehand or adaptively adjusted during the training process. The supervised loss $L^l$ typically refers to the cross-entropy loss computed between the predicted output and the corresponding ${y}_l$, while the unsupervised loss $L^u$ can take on various forms depending on the specific method being used.

The plainest pseudo-label method would be: First, we train the initial model ${M}_0$ on the labeled dataset ${D}_l$ using the cross-entropy loss ${L}^{l}$. This training process generates a pseudo-labeled dataset $\tilde{{D}_u} = {\{({x}_u, M_0(x_u))\}}^q$ for the unlabeled dataset ${D}_u$, where $M_0(x_u)$ represents the pseudo-label of ${x}_u$. 
% Since the size of $q$ is much larger than $p$, this step is crucial for obtaining sufficient training data. 
Next, we combine the labeled dataset ${D}_l$ with the pseudo-labeled dataset $\tilde{{D}_u}$ to form a comprehensive dataset $D = ({D}_l \cup \tilde{{D}_u})$. 
% To ensure that both datasets have the same size, we oversample the labeled dataset ${D}_l$. 
Finally, we train a new model $M$ using the complete dataset $D$.
The aforementioned straightforward procedure can be iteratively performed to consistently improve the quality of the generated pseudo-labels.

%
% \begin{enumerate}
%     \item Firstly, train the model using the labeled data; 
%     \item Assign pseudo-labels to the unlabelled data using the model obtained in 1); 
%     \item Expand the original dataset using the unlabelled data and its corresponding pseudo-labels;
%     \item Re-train the model using the expanded dataset and recycle the above steps until reaching the end condition.
% \end{enumerate}

The process can be significantly influenced by the generation, selection, and improvement of the pseudo-labels due to the unexpected distribution gap and the unsatisfactory performance of pre-trained models.

% 数据集
\subsubsection{Datasets}
Table ~\ref{tab:t0} presents an overview of several commonly utilized datasets that find application in various scenarios. Typically, those fully annotated images are partially selected for semi-supervised learning using ratios such as $5\%, 10\%$, and so on.
% In this section, we present two widely adopted datasets for evaluating semantic segmentation tasks, including PASCAL VOC and Cityscapes.
% \begin{itemize}
% \item \textbf{PASCAL VOC~\cite{everingham2015pascal}} is the most commonly used dataset in semantic segmentation and contains a variety of common scenes and objects.
% % 不要这么详细的讲数据集，简单介绍一下
% \item \textbf{MS-COCO~\cite{lin2014microsoft}.} MS-COCO dataset is the largest dataset with semantic segmentation, providing 80 categories, over 330,000 images, 200,000 of which are labeled, and more than 1.5 million individuals.

% \item \textbf{Cityscapes~\cite{cordts2016cityscapes}} is a dataset containing a large number of street view images, which is widely used in semantic segmentation tasks, and contains more than 19 semantic classes such as roads and vehicles.

% \item \textbf{ADE20K~\cite{zhou2017scene}.} The ADE20K dataset provides a wide range of annotations for scenes, objects, and object parts. It consists of 25,000 images of complex everyday scenes. The training set includes 20,210 images, the validation set includes 2,000 images, and the test set includes 3,000 images.

% \end{itemize}

\begin{table}[htbp]
\begin{tabular}{c|c}
\toprule
\textbf{Category}             & \textbf{Datasets}    \\              
\midrule
\begin{tabular}[c]{@{}c@{}}Natural \\ Images\end{tabular}     & \begin{tabular}[c]{@{}c@{}}
PASCAL VOC\cite{everingham2015pascal},\\ 
MS-COCO\cite{lin2014microsoft},\\
ADE20K\cite{zhou2017scene}
\end{tabular}                       
\\
\hline
\begin{tabular}[c]{@{}c@{}}Street-view \\ Images\end{tabular} & \begin{tabular}[c]{@{}c@{}}
KITTI\cite{geiger2012we},\\
Cityscapes\cite{cordts2016cityscapes} 
\end{tabular}  
\\ \hline
\begin{tabular}[c]{@{}c@{}}Medical \\ Images\end{tabular}      & \begin{tabular}[c]{@{}c@{}}
BRATS\cite{menze2014multimodal},\\
% ACDC\cite{bernard2018deep},\\ 
Kvasir-SEG\cite{jha2020kvasir},\\
LA\cite{xiong2021global}
\end{tabular} 
\\ \hline
\begin{tabular}[c]{@{}c@{}}Satellite \\ Images\end{tabular}   & \begin{tabular}[c]{@{}c@{}}
iSAID\cite{waqas2019isaid},\\
xBD\cite{gupta2019xbd},\\
GID\cite{tong2020land}
\end{tabular}      
\\
\bottomrule
\end{tabular}
\caption{A compilation of the frequently utilized datasets in the domain of semi-supervised semantic segmentation.}
\label{tab:t0}
\end{table}

% 评估标准
\subsubsection{Performance Metrics}
\begin{itemize}
    \item \textbf{Pixel Accuracy} calculates the ratio of correctly classified pixels to the total number of pixels. Although this metric is simple and intuitive, it may not accurately reflect the model's performance when there is an imbalance in the categories.
    \item \textbf{Mean Accuracy} takes into account the pixel precision for each category and calculates the average to address the issue of class imbalance.
    \item \textbf{Mean IoU} calculates the average Intersection over Union $(mIoU)$ between the prediction and ground truth for all categories:
    \begin{align}
    mIoU=\frac{1}{N}\sum_{i=1}^{N}\frac{N_{ii}}{\sum_{j=1}^{N}N_{ij}+\sum_{j=1}^{N}N_{ji}-N_{ii}}
    \end{align}
    where $N$ is the number of categories, $N_{ii}$ is of $TP$(True Positive) number for category $i$, $N_{ij}$ is $FP$(False Positive) number for categories $i$ as $j$, and $N_{ji}$ is $FN$(False Negative) number for categories $j$ as $i$.
    \item \textbf{Weighted IoU} is used as a modification to the mIoU metric in situations where certain categories need to be given more importance.
    % is the total pixels per category over the weighted IoU, where the weights are the number of pixels per category in the real label.
\end{itemize}

\begin{figure*}[!t]
    \centering
    \includegraphics[width=0.85\textwidth]{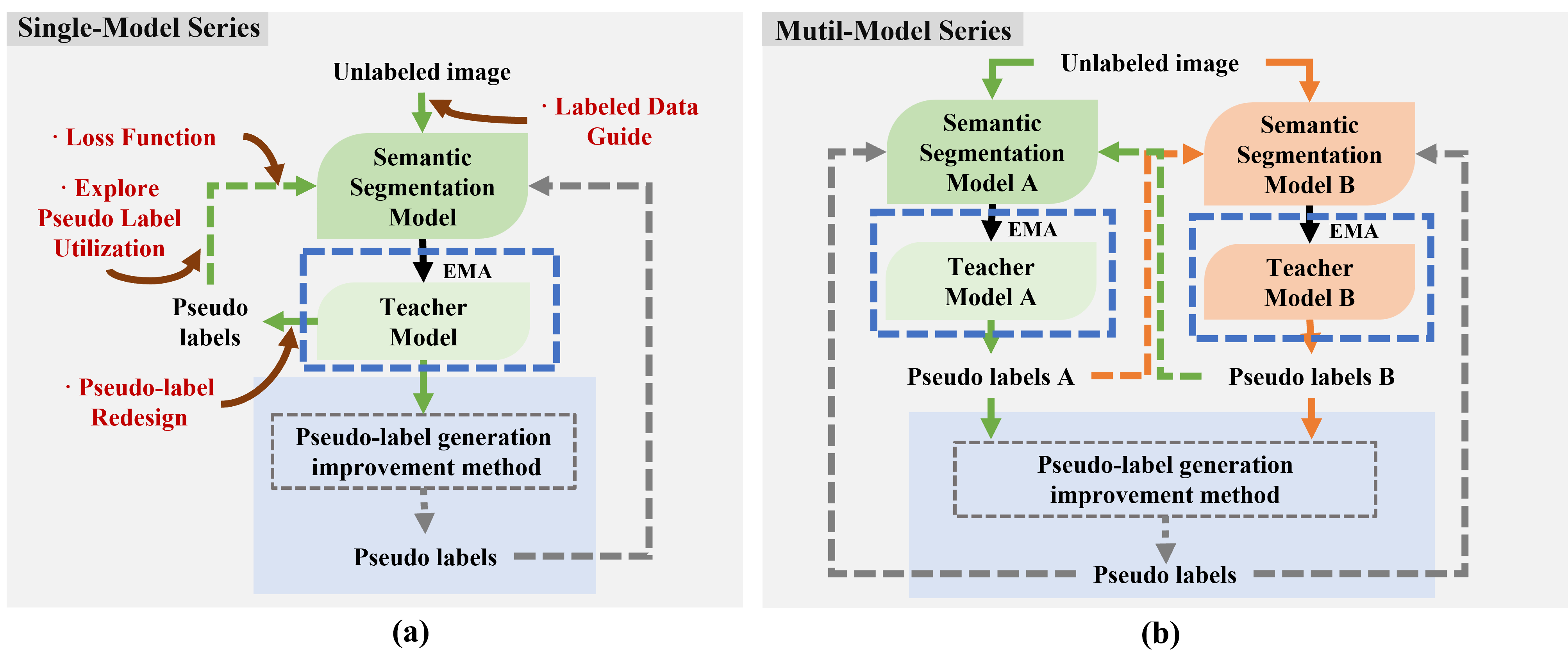}
    \caption{A summary of the primary classifications. Our review is organized into three primary categories: the model for generation (subfigures a and b), the enhancement of pseudo-labels (blue boxed), and the perspective of optimization (dark red arrows). Within the category of the pseudo-labels generation, we investigate two approaches: single-model and multi-model series. Furthermore, various strategies are employed for the selection or refinement of pseudo-labels. Lastly, optimization methods are also becoming more prevalent in this field.}
    \label{fig_meth}
\end{figure*}

% \subsection{!Pseudo-labeling method} \label{2C}

% 我们的分类依据 单独开一章
% 根据图2内容修改
\section{Categorization}
% 点一下现有研究
Drawing on the categorization of network structures for pseudo-label method in previous research~\cite{pelaez2023survey}, as well as the innovative refinement of the Mean Teacher structure proposed by researchers~\cite{tarvainen2017mean}, our investigation will focus on three main areas: the model-based perspective, pseudo-labels refinement, and optimization measures. 
Figure~\ref{fig_meth} presents a comprehensive overview of the various types of pseudo-label techniques.

\noindent \textbf{Model perspective}. Essentially, the various methods for generating pseudo-labels can be divided into two categories: the single-model family and the mutual-model family. Figure~\ref{fig_meth} (a) illustrates the single-model-based approach, where pseudo-labels generated by a single model are used for supervised subsequent training. For example, in the Mean-Teacher method, a single model is trained using pseudo-labels generated by the teacher model, which incorporates consistency regularization. On the other hand, multi-model mutual-training-based approaches aim to improve model performance by jointly training multiple models. Figure~\ref{fig_meth} (b) depicts this approach, where two different networks are initialized and one model supervises the training of the other model by providing pseudo-labels on unlabeled data. The cross-supervision of the two models helps localize and minimize errors in the pseudo-labels.
% 
% 一致性正则化
% On the other hand, the dual model has a teacher model added to each of the two models. 

% The provided figure illustrates two different architectures. The first one is a single-model architecture, where the teacher model generates pseudo-labels. The second one is a two-model architecture, where Model A uses the teacher model of Model B to provide pseudo-supervision. These architectures help to avoid possible interactions between the direct supervision of the two networks and improve the overall effectiveness of the modeling process.

% 伪标签refinement方法
\noindent \textbf{Refinement of pseudo-labels}. We will discuss enhancements to the conventional architecture by focusing on refining pseudo-labels to generate specific labels. Furthermore, we will categorize these methods of refining pseudo-labels into two groups based on whether the pseudo-labels are changed: label updating and filtering only. The simplified architecture for pseudo-labels refinement is depicted in the blue areas of Figure~\ref{fig_meth} (a) and (b). 
% Specifically, CISC-R\cite{wu2023querying} utilizes labeled images to correct inaccurate pseudo labels. On the other hand, PGCL\cite{kong2023pruning} proposes resolving the ambiguity of confidence scores through a network pruning guide to improve the filtering process and obtain high-quality pseudo labels for subsequent training.

\noindent \textbf{Optimization.} Lastly, we also investigated a few emerging optimization techniques, which are depicted in dark red in Figure~\ref{fig_meth} (a). Further elaboration on these methods will be provided in the following sections.

% 正文
% 分成模型和伪标签两部分
\section{Pseudo-Label for Semi-supervised Semantic Segmentation}
In this section, we provide a summary of all the methods for semi-supervised semantic segmentation based on pseudo-labels mentioned in Table \ref{tab:t1}. 
% These methods will be further explained in detail in the following subsections.

\subsection{Model perspective}
The significance of architecture in deep learning cannot be overstated as it establishes the framework and layout of the neural networks employed in these models. The architecture plays a crucial role in determining how input features are manipulated and analyzed by the network, thereby impacting the model's capacity to learn and generate precise predictions. Selecting an appropriate architecture is a vital aspect of constructing a successful deep-learning model. In this subsection, we examine papers pertaining to single-model families and collaborative mutual-model families.

\subsubsection{Single-Model-Based Methods}

%简要介绍，引入
The initial proposal of the single-model self-training approach was made by Yarowsky \cite{yarowsky1995unsupervised}. 
Later on, \cite{lee2013pseudo} suggested combining self-training with neural networks using pseudo-labels. Since the initial single-model self-training approach was straightforward, subsequent research has focused on enhancing different aspects of this network structure.

%自训练过程迭代
\noindent\textbf{Self-training Iteration.}
% From the iterative point of view of the self-training process, 
GIST \& RIST \cite{teh2022gist}, which can be explained as a greedy algorithmic strategy (GIST) and a follow-along iterative self-training strategy-based strategy (RIST) that alternates between ground truth and pseudo-labels.
% 
%ST++
Another approach proposed from a self-training iterative perspective is ST++ \cite{yang2022st++}, where the key step is selective retraining during the iteration process. Since segmentation performance is positively correlated with the evolutionary stability of the generated pseudo-labels during the supervised training phase, more reliable unlabeled images can be selected by evolution during the training process.
The stability metric is based on the average IOU between each early pseudo-mask and the final mask:
\begin{equation}
s_{i}=\sum_{j=1}^{K-1}\operatorname*{mean}\operatorname{IOU}\left(M_{i j},M_{i K}\right)
\end{equation}
The stability scores of all unlabeled images are then obtained and the entire unlabeled set is ranked accordingly, and the R images with the highest scores are selected for the first stage of retraining.
% 
% ST++的选择性
ST++ predicts robustness based on asymptotic changes in reliability, thus eliminating the need to select confidence thresholds for pixel-by-pixel filtering manually.
%, thereby reducing computational costs.
% 列表总结方法
\begin{table*}[!h]
\begin{tabular}{ccc|c|c|c}
\toprule[1pt]
\multicolumn{3}{c|}{\textbf{Category}} &
  \makebox[0.1\textwidth][c]{\textbf{Method}} &
  \textbf{Publication} &
  \textbf{Main contributions} \\ 
  \midrule
\multicolumn{1}{c|}{\multirow{12}{*}
{\rotatebox{90}{\textbf{Model perspective}}}} 
&
  \multicolumn{1}{c|}{\multirow{6}{*}
  {\rotatebox{90}{\textbf{Single Model}}}} &
  \textbf{Self-cross Sup.} &
  USCS\cite{zhang2022semi} &
  ACCV &
  Uncertainty leads self-cross Sup. \\ \cline{3-6} 
\multicolumn{1}{c|}{} &
  \multicolumn{1}{c|}{} &
  \multirow{2}{*}{\textbf{\begin{tabular}[c]{@{}c@{}}Self-training\\ Iteration\end{tabular}}} &
  GIST\& RIST\shortcite{teh2022gist} &
  CRV &
  Greedy and Randomized Iterative Self-Training \\ \cline{4-6} 
\multicolumn{1}{c|}{} &
  \multicolumn{1}{c|}{} &
   &
  ST++\cite{yang2022st++} &
  CVPR &
  \multicolumn{1}{c}{Select reliable images for retraining} \\ \cline{3-6} 
\multicolumn{1}{c|}{} &
  \multicolumn{1}{c|}{} &
  \multirow{3}{*}{\textbf{\begin{tabular}[c]{@{}c@{}}Auxiliary\\ Tasks or\\Network\end{tabular}}} &
  \cite{li2021residual} &
  PRCV &
  \multicolumn{1}{c}{Residual correction approach} \\ \cline{4-6} 
\multicolumn{1}{c|}{} &
  \multicolumn{1}{c|}{} &
   &
  ELN\cite{kwon2022semi} &
  CVPR &
  \multicolumn{1}{c}{\begin{tabular}[c]{@{}c@{}}Auxiliary Modules:Error localization network,\\ ELN training with constrained loss function\end{tabular}} \\ \cline{4-6} 
\multicolumn{1}{c|}{} &
  \multicolumn{1}{c|}{} &
   &
  EPS++\cite{lee2023saliency} &
  TPMAI &
  \multicolumn{1}{c}{\begin{tabular}[c]{@{}c@{}}Explicit pseudo-pixel supervision,\\ Using saliency as pseudo-pixel supervision\end{tabular}} \\ 
  % \cline{2-6} 
  \cmidrule{2-6}
\multicolumn{1}{c|}{} &
  \multicolumn{1}{c|}{\multirow{6}{*} 
  {\rotatebox{90}{\textbf{Mutual Model}}}} &
  \multirow{4}{*}{\textbf{\begin{tabular}[c]{@{}c@{}}Cross\\ Pseudo\\ Sup.\end{tabular}}} &
  CPS\cite{chen2021semi} &
  CVPR &
  \multicolumn{1}{c}{Cross-pseudo-supervision} \\ \cline{4-6} 
\multicolumn{1}{c|}{} &
  \multicolumn{1}{c|}{} &
   &
  n-CPS\cite{filipiak2021n} &
  ArXiv &
  \multicolumn{1}{c}{N networks cross-pseudo supervision} \\ \cline{4-6} 
\multicolumn{1}{c|}{} &
  \multicolumn{1}{c|}{} &
   &
  UCC\cite{fan2022ucc} &
  CVPR &
  \multicolumn{1}{c}{Uncertainty-guided cross-head co-training} \\ \cline{4-6} 
\multicolumn{1}{c|}{} &
  \multicolumn{1}{c|}{} &
   &
  CCVC\cite{wang2023conflict} &
  CVPR &
  \multicolumn{1}{c}{\begin{tabular}[c]{@{}c@{}}Conflict cross-view consistency,\\ Co-training two network branches\end{tabular}} \\ \cline{3-6} 
\multicolumn{1}{c|}{} &
  \multicolumn{1}{c|}{} &
  \multirow{2}{*}{\textbf{\begin{tabular}[c]{@{}c@{}}Dynamic\\ Muti-train\end{tabular}}} &
  DMT \cite{feng2022dmt} &
  PR &
  \multicolumn{1}{c}{Dynamic mutual training} \\ \cline{4-6} 
\multicolumn{1}{c|}{} &
  \multicolumn{1}{c|}{} &
   &
  DMT-PLE \cite{ke2022three} &
  Access &
  \multicolumn{1}{c}{Pseudo-label enhancement strategy} \\ 
  \midrule
\multicolumn{1}{c|}{\multirow{11}{*}
{\rotatebox{90}{\textbf{Pseudo-label Refinement}}}}
& \multicolumn{1}{c|}{\multirow{6}{*}
  {\rotatebox{90}{\textbf{Label Update}}}}
  &\multirow{4}{*}{\textbf{\begin{tabular}[c]{@{}c@{}}Pseudo-label \\ Corrections\end{tabular}}} &
  \cite{yi2021learning} &
  TIP &
  \multicolumn{1}{c}{Graph-based noise labels correction} \\ \cline{4-6} 
\multicolumn{1}{c|}{} &
  \multicolumn{1}{c|}{} &
   &
  CARD\cite{wang2022card} &
  IJCAI &
  \multicolumn{1}{c}{Semantic linking to correct noisy labels} \\ \cline{4-6} 
\multicolumn{1}{c|}{} &
  \multicolumn{1}{c|}{} &
   &
  ThreeStageSelftraining  \shortcite{ke2022three} &
  TIP &
  \multicolumn{1}{c}{Multi-task count and update pseudo-labels} \\ \cline{4-6} 
\multicolumn{1}{c|}{} &
  \multicolumn{1}{c|}{} &
   &
  CISC-R\cite{wu2023querying} &
  TPAMI &
  \multicolumn{1}{c}{Labeled images correct inaccurate pseudo-label} \\ \cline{3-6} 
\multicolumn{1}{c|}{} &
  \multicolumn{1}{c|}{} &
  \multirow{2}{*}{\textbf{De-biasing}} &
  DARS\cite{he2021re} &
  ICCV &
  \multicolumn{1}{c}{
  \begin{tabular}[c]{@{}c@{}}Redistribute biased pseudo-labels,\\ aligned with the true distribution
  \end{tabular}} \\ \cline{4-6} 
\multicolumn{1}{c|}{} &
  \multicolumn{1}{c|}{} &
   &
  DST\cite{chen2022debiased} &
  NeurIPS &
  \multicolumn{1}{c}{De-biased self-training} \\ 
  \cmidrule{2-6} 
\multicolumn{1}{c|}{} &
  \multicolumn{1}{c|}{\multirow{5}{*}
  {\rotatebox{90}{\textbf{Filter-only}}}}
  &\multirow{3}{*}{\textbf{\begin{tabular}[c]{@{}c@{}}Confidence \\ Filtering\end{tabular}}} &
  C3-SemiSeg\shortcite{zhou2021c3} &
  ICCV &
  \multicolumn{1}{c}{Dynamic Confidence Region Selection Strategy} \\ \cline{4-6} 
\multicolumn{1}{c|}{} &
  \multicolumn{1}{c|}{} &
   &
  CAFS\cite{ju2023cafs} &
  ArXiv &
  \multicolumn{1}{c}{Adaptive classification confidence thresholds} \\ \cline{4-6} 
\multicolumn{1}{c|}{} &
  \multicolumn{1}{c|}{} &
   &
  TorchSemiSeg2\shortcite{chen2023semi} &
  ICME &
  \multicolumn{1}{c}{Local Pseudo Label Filtering Module} \\ \cline{3-6} 
\multicolumn{1}{c|}{} &
  \multicolumn{1}{c|}{} &
  \textbf{\begin{tabular}[c]{@{}c@{}}Confidence \\ Refinement\end{tabular}} &
  PGCL\cite{kong2023pruning} &
  WACV &
  \multicolumn{1}{c}{Network Pruning Refinement Confidence Score} \\ \cline{3-6} 
\multicolumn{1}{c|}{} &
  \multicolumn{1}{c|}{} &
  \textbf{\begin{tabular}[c]{@{}c@{}}Assisted Net \\ Filtering\end{tabular}} &
  GTA-Seg\cite{jin2022semi} &
  NeurIPS &
  \multicolumn{1}{c}{Assistants Teacher sift useful information} \\
  \midrule
\multicolumn{1}{c|}{\multirow{5}{*}
{\rotatebox{90}{\textbf{Optimization}}}}
&  \multicolumn{2}{c|}{\multirow{2}{*}{\textbf{\begin{tabular}[c]{@{}c@{}}Loss \\ Function\end{tabular}}}} &
  \cite{wang2022learning} &
  PR &
  \multicolumn{1}{c}{Class-aware Cross-entropy loss} \\ \cline{4-6} 
\multicolumn{1}{c|}{} &
  \multicolumn{2}{c|}{} &
  PS-MT\cite{liu2022perturbed} &
  CVPR &
  \multicolumn{1}{c}{Confidence-weighted Cross-entropy(Conf-CE)} \\ \cline{2-6} 
\multicolumn{1}{c|}{} &
  \multicolumn{2}{c|}{\textbf{\begin{tabular}[c]{@{}c@{}}Labeled data \\ Utilization\end{tabular}}} &
  GuidedMix-Net\shortcite{tu2022guidedmix} &
  AAAI &
  \multicolumn{1}{c}{Label information guide unlabeled learning} \\ \cline{2-6} 
\multicolumn{1}{c|}{} &
  \multicolumn{2}{c|}{\textbf{\begin{tabular}[c]{@{}c@{}}Pseudo-label \\ Redesign\end{tabular}}} &
  PseudoSeg\shortcite{zou2020pseudoseg} &
  ICLR &
  \multicolumn{1}{c}{Structuring and quality redesign pseudo-label} \\ \cline{2-6} 
\multicolumn{1}{c|}{} &
  \multicolumn{2}{c|}{\textbf{\begin{tabular}[c]{@{}c@{}}Pseudo-label \\ Tradeoffs\end{tabular}}} &
  CPCL\cite{fan2023conservative} &
  TIP &
  \multicolumn{1}{c}{Conservative\&progressive explore pseudo-label} \\ 
  \bottomrule
\end{tabular}
\caption{An overview of pseudo-label techniques in the field of semi-supervised semantic segmentation.}
\label{tab:t1}
\end{table*}

%自交叉监督
\noindent\textbf{Self-cross Supervision.} \cite{zhang2022semi} proposes a method called uncertainty-guided self-cross-supervision (USCS) for semi-supervised semantic segmentation. This approach utilizes the results of a multiple-input multiple-output (MIMO) segmentation model to perform self-cross-supervision, resulting in significant reductions in parameter and computation costs.

% 辅助结构
\noindent\textbf{Auxiliary Tasks Framework.}
Since another key to the pseudo-label approach is to set up auxiliary task frame supervision, some approaches have been proposed from this perspective.
%残差辅助网络
Earlier, \cite{li2021residual} introduced residual networks to extend the self-training structure. The labeled data is fed into an auxiliary residual network to predict the residuals from the original segmentation results.
% ELN
While the later proposed ELN \cite{kwon2022semi} is mainly to assist in locating errors, the auxiliary module is trained to identify the pixel points that may be incorrect by taking the images and segmentation results as inputs. 
The specific ELN structure contains the main segmentation network (encoder and decoder) and the auxiliary decoder (\textit{$D_1$, $D_2$, \dots, $D_K$}). 
The main segmentation network is trained by standard cross-entropy loss, while the auxiliary decoder is trained by restricted cross-entropy loss:
% 换行很奇怪，修改一下表达
\begin{align}
\displaystyle{\cal L}_{a u x} & =  \frac{1}{|D_{L}|} \sum_{X\in D_{L}} \sum_{k=1}^{K} 
\notag  \\ 
& \left\{L_{ce}\left (P^{k},Y \right ) >\alpha^{k} \cdot  L_{ce}\left (P,Y \right )  \right\} \cdot L_{ce}\left (P^{k},Y \right )
\end{align} \label{eq1}
As a result, the auxiliary decoder will perform much worse than the primary decoder because it contains various errors, which are then used as inputs to the ELN to train the ELN to locate the labeling errors, similar to manually creating some error data for training.
% 
% 显著性作为伪标签
EPS++\cite{lee2023saliency} provides rich boundaries through the saliency map generated by the saliency detection model, which is combined with image-level labeling information for joint training, assisting the model in being trained from pixel-level feedback.

% \begin{figure}[!t]
%     \centering
%     \includegraphics[width=1\linewidth]{fig/fig_2.png}
%     \caption{The two-model mutual-learning network structure operation (adapted from DMT) trained two different models on a labeled subset, with one model providing pseudo-supervision for the other. Where the visual samples are selected from the PASCAL VOC dataset.}
%     \label{fig_mnet}
%     % \setlength{\floatsep}{2pt plus 2pt minus 2pt}
% \end{figure}

\subsubsection{Mutil-model-based Methods}
Several semi-supervised learning methods rely on pseudo-supervision, particularly self-training methods that generate pseudo-labels. However, the pseudo-labels generated by a single model in self-training can often be unreliable. This is because it is common to use only a single model's prediction confidence to filter out low-confidence pseudo-labels, which can leave behind high-confidence errors and waste many low-confidence correct labels.
The dual-model mutual-training method aims at the problems inherent in the single-model self-training method, i.e., a single model is unable to detect and correct its errors, which may result in the accumulation of bias and ultimately affect the training and segmentation effects, 
mutual-training \cite{zhang2018deep} 
by which two or more models train each other according to their differences, localize their errors, and correct each other is proposed, 
% Wang et al.\shortcite{wang2010new} show in a study that the co-training process is viewed as combined label propagation on two views, which provides a feasible solution for incorporating graph-based and divergence-based semi-supervised methods into a unified framework.
% Similar approaches in semi-supervised techniques other than pseudo-label have been categorized as co-training, multi-view constraints, etc.

%交叉监督
\noindent\textbf{Cross-Pseudo-Supervision.}
A classical inter-training perspective is dual-model cross-supervision, such as CPS~\cite{chen2021semi} approach uses different initialization methods for two networks, where the pseudo-labels output by one network supervise the other segmentation network. 
The subsequently proposed n-CPS~\cite{filipiak2021n} is the result of extending the CPS to n sub-networks, and experiments demonstrate that network integration significantly improves performance.
% TorchSemiSeg2 \cite{chen2023semi}, similarly employs a cross-pseudo-supervision strategy, where images are fed into two parallel segmentation networks, and the pseudo-labels generated by the networks supervise each other. 
% 
% UCC
Based on cross-supervision, ~\cite{fan2022ucc} introduces uncertainty-guided supervision and proposes UCC (Uncertainty-guided Cross-head Co-training of cross-heads), which further improves the generalization ability by sharing encoders.
% A method called DCSCP is proposed to increase the diversity of consistent training samples while reducing distributional inconsistencies and addressing class imbalance. In addition, we propose UGRM address the pseudo-labeling noise introduced by self-training
% 
%CCVC 详细写
\cite{wang2023conflict} design a conflict-based cross-try consistency (CCVC) to force two subnets to learn knowledge from unrelated views.
%CVC详细
They propose a new cross-view consistency strategy to encourage two subnets, which are structurally similar but do not share parameters, to learn different features from the same input image. They introduce a feature difference loss to achieve this. 
For unlabeled data, they make the two subnets use each other's pseudo-labels for model learning.
% 
%CPL 详细 
% Similarly, to ensure that the model learns more useful information from conflicting predictions, they also refer to a previous method \cite{zhang2018collaborative},
% %yang2023revisiting
% Conflicting predictions are further categorized into two types: conflicting and credible and conflicting but incredible.

% 动态互训练
\noindent\textbf{Dynamic Muti-training.}
In addition to cross-supervision for synchronous training, other researchers have proposed dynamic mutual training based on a two-network structure, where two networks are trained asynchronously.
%DMT详细写
DMT \cite{feng2022dmt} points out that it is difficult for a single model to overcome its own errors. Therefore, they use two models with different initializations, one of which generates offline pseudo-labels for the other proposed.
To effectively train machine learning models, it is important to identify labeling errors. This they do by comparing the predictions of two different models and quantifying the differences between them, so that they can dynamically adjust the weight loss during training to improve the accuracy of the model.
Dynamic loss weight ${\omega}_{u}$ is defined as follows:
\begin{equation}
    {\omega}_{u}=
    \begin{cases}
    p_{B}^{\gamma_{1}}, & y_{A}=y_{B} \\
    p_{B}^{\gamma_{2}}, & y_{A}\neq y_{B}, c_{A}\geq c_{B} \\
    0,  & y_{A}\neq y_{B}, c_{A}<c_{B}
    \end{cases}
\end{equation}
Regarding the issue of noise in the pseudo-label technology, the DMT method proposes assigning different weights to the samples, rather than discarding them. This approach is intended to preserve low-confidence data, but it may not be effective in addressing the problem of high noise rates in the pseudo-label method. Additionally, the "catastrophic forgetting problem" in neural networks cannot be fully resolved, since the basic components of these networks consist of fixed structures and parameters. However, the severity of this problem can be reduced through various mitigation techniques.
%\cite{mccloskey1989catastrophic}.
% DMT-PLE
The DMT-PLE\cite{zhou2022catastrophic}method extends the pseudo-label enhancement strategy from the previous DMT method, mainly for the purpose mentioned above. 
They mention that it's challenging for a model to retain the knowledge it has learned when processing input with multiple pixels of varying types. 
To prevent the model from developing a bias towards the last learned category, they use a strategy called Pseudo-Label Enhancement (PLE). 
This technique utilizes the pseudo-labels generated by the model in the previous stage to refine pseudo-labels generated by the current model.

\subsection{Pseudo-label Refinement Methods}
\subsubsection{Pseudo-label Update Methods}
The pseudo-label method sometimes leads to incorrect predictions or inaccurate pseudo-labels being included in the training process, which can accumulate errors and make the learned pseudo-labels ineffective in guiding the subsequent learning, and ultimately affect the training results of the segmentation model. To solve this problem, some studies have proposed methods for updating pseudo-labels to alleviate the noise problem with good results.

\noindent\textbf{Pseudo-label Corrections. } 
%噪声标签学习角度
In the beginning, some works formulated the task as a learning problem of pixel-level label noise.
Yi et al.~\shortcite{yi2021learning} introduce a graph-based label noise detection and correction framework, which utilizes pixel-level labels generated by class activation maps (CAMs) as weakly annotated noise labels, trains a strongly annotated segmentation model to detect clean labels from the above noisy labels, and then corrects the noisy labels using a clean label supervised graph attention (GAT) network.
The clean label supervised graph attention network is then used to correct the noise labels.
% 语义连接纠正 CARD
Similarly, to address the problem of noisy label correction, ~\cite{wang2022card} propose a category-independent relational network to correct labels based on reliable semantic associations between image features.
They obtained relational estimates by augmenting the relationships between features. The predictions with weak correlations are discarded for effective noise label correction.

% 三阶段训练
Unlike the perspectives of the above approaches, some of the improvement methods start with the training phase. 
Ke et al.\shortcite{ke2022three} advance a ThreeStageSelftraining method, where they attempted to extract initial pseudo-labels information on unlabeled data through three stages of self-training while enforcing segmentation consistency in a multitasking manner to generate higher quality pseudo-labels.
%提一下DMT-PLE的伪标签增强
% Similarly, DMT-PLE provides a pseudo-labeling enhancement strategy, i.e., the pseudo-labeling in the current phase is mainly updated by retaining and introducing into the model training the a priori rehearsal knowledge saved in the last iteration of the previous phase.
% 
% \begin{figure}[!t]
%     %\centering
%     \includegraphics[width=1\linewidth]{fig/CISC-R image select.png}
%     \caption{CISC-R\cite{wu2023querying} proposed image selection method. Darker colored circles indicate images with smaller D-values.}
%     \label{fig_ciscRis}
% \end{figure} 
% 
%详细写CISC-R
In addition to the traditional remedies described above, ~\cite{wu2023querying} raise the CISC-R method using labeled images to correct noisy pseudo-labels, considering that samples of the same kind have a high pixel-level correspondence. 
Inspired by ST++~\cite{yang2022st++}, they used a CISC-based image selection method that takes into account inter-class feature differences and the difficulty of correcting noisy pseudo-labels at the beginning of training.
First, 
% as shown in Figure. ~\ref{fig_ciscRis}, 
for each class $k$, an initial model is used to extract the anchor vector $a^{k}_{l}$ for that class from the set of labeled images:
\begin{align}
    a_l^k=\frac{1}{n_l^k}\sum_i^{n_l^k}v_l^k=\frac{1}{n_l^k}\sum_i^{n_l^k}F_l^k\odot m_l^k .
\end{align}
Through this average generation, $a^{k}_{l}$ is generated to represent the categorized anchor points of the labeled images.
Specifically, a CISC mapping $m'$ is generated by cosine similarity between $a^{k}_{l}$ of the labeled image x and the high-level feature. 
% Related research addressing noise labeling is not limited to single-model self-training, and some work has also utilized the idea of multi-model fusion for mutual training ensemble improvement.
% Since heterogeneous network architectures are important for mitigating model coupling, Zhang et al.\cite{zhang2021robust} devised a robust mutual learning approach that introduces an average teacher model for each model based on two-model mutual training, such that the two models supervise each other through each other's teacher models without direct connectivity interactions, and secondly, the approach also utilizes the network knowledge to correct noisy pseudo-labeling.

\noindent\textbf{De-biasing}  
Bias in training comes both from the network itself and from improper training of potentially incorrect pseudo-labels, which accumulate errors throughout iterations.
% DARS
\cite{he2021re} propose Distributed Alignment and Random Sampling (DARS), a simple and effective method to redistribute biased pseudo-labels, to align pseudo-labels with the ground truth, and to improve the effect of noisy labels on training. Aligning their distribution with the true distribution improves semi-supervised semantic segmentation.
% 去偏差自训练 DST
Subsequently, to minimize the bias, Chen et al.\shortcite{chen2022debiased} suggest the debiased self-training (DST). The key of this method is that two parameter-independent classifier headers decouple the process of generating and utilizing pseudo-labels, and only clean labels are used for training, which improves the quality of pseudo-labels.

\subsubsection{Pseudo-label Filtering-only Methonds}
In addition, some researchers have proposed to enhance the segmentation effect by filtering the noisy pseudo-labels, and such methods do not perform the update of the noisy labels.

\noindent\textbf{Confidence  Filtering.}
Zhou et al.\shortcite{zhou2021c3} propose C3-SemiSeg, which presents a dynamic confidence region selection strategy to focus on high-confidence regions for loss computation. In addition, cross-set contrast learning is also integrated to improve feature representation.
% cafs
However, to address the problem that existing high-confidence-based pseudo-label methods lose most of the information, Ju et al.\shortcite{ju2023cafs} propose a class-adaptive semi-supervised framework (CAFS) for semi-supervised semantic segmentation, which allows the construction of a validation set on the labeled dataset to take advantage of the calibration performance of each class. This includes a core operation: adaptive class-by-class confidence thresholding (ACT), which de-emphasizes the use of calibration scores to adaptively adjust reliability confidence thresholds.
% 
%点一下前文提到的Torch方法
Recently, TorchSemiSeg2\cite{chen2023semi} introduces a localized pseudo-labels filtering module to assess the reliability of region-level pseudo-labels using a discriminator network. They also propose a dynamic region loss correction to further assess the reliability of pseudo-labels using network diversity and to evaluate the direction of convergence of the network.

\noindent\textbf{Confidence-level Refinement. }
%最近研究
Since much of the previous work mentioned above evaluated pseudo-label data mostly based on confidence thresholds, the problem of confidence ambiguity that exists at the beginning of training may largely limit subsequent updates.
Recently, Kong et al. introduced the PGCL~\shortcite{kong2023pruning}, which aims to solve the issue of fuzzy confidence scores in network pruning, 
% . They used a new method to create better-quality pseudo-labels 
by gradually teaching the network from easy to hard examples using a coarse strategy.

\noindent\textbf{Assisted Net Filtering. }
In addition, some methods utilize auxiliary structures for filtering operations. For example, GTA-Seg\cite{jin2022semi} selects an auxiliary structure known as the gentle Teaching Assistant. The GTA learns directly from the pseudo-labels generated by the teacher's network, and only filtered, favorable information is passed on to the student's network to assist in supervising the training of the student's network.

\subsection{Optimization Methods}
In addition to the above methods, some researchers have proposed unique optimization techniques to improve segmentation results. These techniques include loss function improvement, pseudo-labels redesign, etc.

% \subsubsection{Loss Function}
\noindent\textbf{Loss Function.}
% Learning pseudo labels
Wang et al. \shortcite{wang2022learning} advise improving the quality of pseudo-labels through perceptual cross-entropy (CCE) and progressive cross-training (PCT). CCE can simplify the pseudo-label generation more than traditional cross-entropy. PCT gradually introduces high-quality predictions as additional supervision for network training.
% PS-MT
PS-MT\cite{liu2022perturbed} uses a stricter confidence-weighted cross-entropy (Conf-CE) to address the problem that cross-entropy loss training can easily overfit prediction errors.

% \subsubsection{Labeled data Utilization}
\noindent\textbf{Labeled data Utilization.}
% 标记数据指导
It's worth noting that \cite{tu2022guidedmix} considers that dealing with labeled and unlabeled data separately usually results in discarding a large amount of a priori knowledge learned from labeled examples. So they proposed a method called GuidedMix-Net, which learns higher-quality pseudo-labels by using labeling information to guide the learning of unlabeled examples.
% CISC-R利用标记数据指导纠正伪标签
Similarly, CISC-R\cite{wu2023querying}, which we mentioned earlier, corrects the pixel-level correction of pseudo-labels by estimating the pixel similarity between the unlabeled image and the queried labeled image and generating the CISC map.

% \subsubsection{Pseudo-label Utilization}
\noindent\textbf{Pseudo-label Utilization.}
% 伪标签重新设计
%PseudoSeg详细写
Zou et al.\shortcite{zou2020pseudoseg} focus on structured and qualitative design methods for pseudo-labels and proposed a single-stage consistency training framework, PseudoSeg, which generates pseudo-labels from two branches: the output of the segmentation model and the output of the class activation maps (CAM).
They propose a pseudo-labels redesign strategy that combines pseudo-labels from two sources through a calibrated fusion strategy, i.e., given a batch of decoder outputs $\hat{p} = f_{\theta }(\omega (x))$ and SGC mappings computed based on the weakly augmented data $w(x)$, pseudo-labels $\tilde{y}$ are generated:
\begin{equation}
    \begin{aligned}
    \mathcal{F}(\hat{p},\hat{m})=&\text{Sharpen}(\gamma\text{ Softmax}\left(\frac{\hat{p}}{\text{Norm}(\hat{p},\hat{m})}\right)
    \\ &+(1-\gamma)\text{ Softmax}\left(\frac{\hat{m}}{\text{Norm}(\hat{p},\hat{m})}\right),T)
\end{aligned}
\end{equation}
PseudoSeg generates well-calibrated, high-quality pseudo-labels by implementing a novel pseudo-labels redesign strategy that facilitates subsequent model training.
With limited available labeled data, well-calibrated pseudo-labels can greatly improve segmentation. To further improve the calibration, they suggested exploring advanced techniques such as multimodal data fusion.

% 伪标签权衡利用
\cite{fan2023conservative} develop Conservative Progressive Collaborative Learning (CPCL) from the perspective of dual-model mutual training, where the conservative branch is cross-supervised using high-quality pseudo-labels to achieve conservative protocol-based evolution.
The progressive branch is supervised by a union utilizing a large number of labels to achieve progressive exploration of divergence

% 其他领域的伪标签方法 改一下标题。
% 在前面指出一下前面的方法主要是聚焦于自然图像领域。
\section{Pseudo label Methods in Other Areas}
Pseudo-label method is widely used in semantic segmentation because of its simplicity and effectiveness. The methods we summarized in the previous section are mainly focused on natural image segmentation, but there is no doubt that it is important to continue research and promotion of image segmentation in more fields.
In this section, we focus on pseudo-label techniques applied to some specific domains, including medical images and remote sensing image segmentation.
% In Table~\ref{tab:t2}, we summarize the pseudo-label application methods covered in this paper, including medical images and remote sensing images.

\subsubsection{Medical Image Segmentation}
Segmenting medical images, which involves identifying the pixels corresponding to organs or lesions in images like CT or MRI images, is a highly difficult task in medical image analysis due to a lack of sufficient labels. 
Numerous studies have proposed the use of pseudo-labels, yielding promising outcomes when applied to specific medical datasets.
%单模型
%ASTO
% In the early, \cite{huo2021atso} noticed a weakness in the model known as "inert imitation", whereby the model would tend to retain pre-existing predictions and thus resist updates.
In their recent work, \cite{huo2021atso} introduces a novel approach called asynchronous teacher-student optimization (ATSO) to challenge the conventional learning strategy. Instead of training two models alternately, they propose dividing the unlabeled training data into two subsets. This approach is particularly applied to 3D medical images, where each 3D body is divided into 2D slices representing coronal, sagittal, and axial views. A 2D network is then employed for segmentation, and the resulting output is stacked to form 3D bodies.
% 
%EMSSL
% Whereas \cite{xu2023expectation} proposes a new method EMSSL for improving pseudo-labeling based on the original pseudo-labeling method, i.e., establishing a link between pseudo-labeling and Expectation Maximization(EM)Algorithm, which provides a complete generalization of the generalized pseudo-labeling under the Bayesian principle, and then provides a variational method for learning the Bayesian pseudo-labeling, which increases the interpretability of the model to some extent.
% 
%BCP
A common issue in semi-supervised medical image segmentation is the discrepancy between the distribution of labeled and unlabeled data. Prior studies have primarily addressed labeled and unlabeled data in isolation or with inconsistency, which can result in disregarding the knowledge gained from the labeled data. In their research, \cite{bai2023bidirectional} propose a straightforward approach to alleviate this problem by incorporating labeled and unlabeled data in both directions using a Mean-Teacher model called BCP.
%, which encourages unlabeled data to learn comprehensive common semantics. 

\subsubsection{Remote Sensing Image Segmentation}
The process of labeling high-resolution remote-sensing satellite images is a task that requires a significant amount of time and effort. This limitation affects the performance of segmentation models. To address this issue, certain reports propose the utilization of pseudo-label techniques that rely on semi-supervised learning. These methods aim to assist in the segmentation of remote-sensing images.
In their paper, \cite{li2023semi} propose a technique for enhancing the segmentation accuracy of limited-sample high-resolution remote sensing images. They achieve this by utilizing two networks, namely UNet %\cite{ronneberger2015u} 
and DeepLabV3 %\cite{chen2017rethinking}
, to predict pseudo-labels and filter them effectively.
% to improve the accuracy of semantic segmentation in the case of limited labeled samples.
% 从而在标记样本有限的情况下提高语义分割的准确性。
% 
%26 详细写
In their recent work, \cite{cui2023semi} propose a novel approach that utilizes bicommutative entropy consistency and a teacher-student structure. This task is challenging due to the presence of multiple classes, complex terrain, significant overlap between classes, and indistinct features. To address these challenges, the authors incorporate a channel attention (CA) mechanism into the teacher coding network. This CA module effectively filters the feature mapping and suppresses noise interference, thereby constraining feature extraction and reducing the information entropy generated by the coding network.
% They also developed two student networks that share a coding network and applied a sharpening function to minimize the uncertainty in unsupervised predictions for both networks.
% 他们还开发了两个共享一个编码网络的学生网络，并应用锐化函数来最小化两个网络的无监督预测的不确定性。

% 应用总结
% Based on the above categorization elaboration summary, we know that the most popular application directions for image segmentation are clearly in the field of medical and remote sensing images. This is because these images are usually of high resolution and, as many related algorithms have demonstrated, incorporating a certain amount of prior knowledge is more likely to yield significant results. Because high-resolution images benefit from incorporating prior knowledge in related algorithms. In addition to the lack of large amounts of accurately labeled data, researchers are delving more deeply into semi-supervised models.

\section{Challenges and Future Perspectives}
% 不是特别新的挑战 不是新的点
% 架构 融入大规模预训练模型等等
% 模型学习层次 低置信度标签的使用 更深入的用
% 主动学习 角度
% 其他新型损失
% 伪标签：不同层次伪标签，
Upon conducting a thorough examination, it is evident that pseudo-label techniques have ventured into various techniques for image segmentation, yielding remarkable outcomes. Nevertheless, this section will concentrate on the difficulties encountered in the pseudo-label method for semi-supervised semantic segmentation and emphasize potential research directions.

\noindent\textbf{Quality enhancement using foundation models.} 
Foundation models have transformed AI, powering prominent chatbots and generative AI.
A cutting-edge interactive prompt-based model called Segment Anything Model (SAM)~\cite{kirillov2023segment} has recently been integrated into semantic segmentation tasks. It is anticipated that in the future, the prompt functionality of SAM will be leveraged to further improve the efficiency and effectiveness of the pseudo-label.

\noindent\textbf{Utilization of additional information. } 
At present, the use of low-quality pseudo-labels is limited to a single type of supervised signal, disregarding the valuable information present in other pixels. Hence, there is an opportunity to integrate alternative forms of supervisory signals into the model, enhancing its capacity to effectively utilize both coarse and fine-labeled data. We anticipate that future studies will enhance segmentation performance by adopting a more holistic approach to supervision.

% \noindent\textbf{Methods of the prototype.} 
% The prototype-based semi-supervised approach is an advanced solution proposed for the issue of comparative learning, aimed at enhancing the model's intra-class compactness. The concept of prototyping has been extensively researched in other fields such as object detection, and we believe it holds great potential for further development in the realm of semantic segmentation.

\noindent\textbf{Engage in the active selection and refine the process.} 
Pseudo-label techniques struggle to effectively resolve the problem of noisy data. Instead of training the model on the entire dataset, strategies like active learning involve selecting a subset of the most informative data points to query for additional labels. This approach is more efficient and cost-effective because it allows the model to learn from the most informative examples without requiring the entire dataset to be labeled. The future looks promising when active selection and refinement strategies are incorporated.

\noindent\textbf{Explore complex segmentation scenarios.} Expanding the application of pseudo-label models to a wider range of real-world situations is essential. While there has been notable advancement in theoretical research, the current utilization of pseudo-label methods is limited to specific datasets such as PASCAL VOC 2012~\shortcite{everingham2015pascal}, which consists of only 20 commonly occurring categories. To advance this field, it is crucial to investigate datasets that better represent real-life scenarios. For example, ADE20K~\cite{zhou2017scene} contains over 150 classes of object information and could serve as a more representative dataset for future exploration.

\section{Conclusion}
We are the first to provide a comprehensive overview and categorization of pseudo-label techniques in the realm of semi-supervised semantic segmentation. Our categorization is based on the viewpoint of the model, methods for refining pseudo-labels, and innovative optimization approaches. Furthermore, we have examined various pseudo-label techniques employed in medical and remote-sensing image segmentation. Lastly, we have identified the current obstacles in this domain and proposed potential future directions. We have also put forth research avenues to tackle these challenges.

\appendix

% 伦理声明
% \section*{Ethical Statement}

% There are no ethical issues.

% \section*{Acknowledgments}

% The preparation of these instructions and the \LaTeX{} and Bib\TeX{}
% files that implement them were supported by Schlumberger Palo Alto
% Research, AT\&T Bell Laboratories, and Morgan Kaufmann Publishers.
% Preparation of the Microsoft Word file was supported by IJCAI.  An
% early version of this document was created by Shirley Jowell and Peter
% F. Patel-Schneider.  It was subsequently modified by Jennifer
% Ballentine, Thomas Dean, Bernhard Nebel, Daniel Pagenstecher,
% Kurt Steinkraus, Toby Walsh, Carles Sierra, Marc Pujol-Gonzalez,
% Francisco Cruz-Mencia and Edith Elkind.

% 参考文献及格式
%% The file named. bst is a bibliography-style file for BibTeX 0.99c
\bibliographystyle{named}
\bibliography{myBib}

\end{document}